\def\year{2024}\relax
\documentclass[letterpaper]{article} % DO NOT CHANGE THIS
\usepackage{aaai24}  % DO NOT CHANGE THIS
\usepackage{times}  % DO NOT CHANGE THIS
\usepackage{helvet}  % DO NOT CHANGE THIS
\usepackage{courier}  % DO NOT CHANGE THIS
\usepackage[hyphens]{url}  % DO NOT CHANGE THIS
\usepackage{graphicx} % DO NOT CHANGE THIS
\usepackage{natbib}  % DO NOT CHANGE THIS AND DO NOT ADD ANY OPTIONS TO IT
\usepackage{caption} % DO NOT CHANGE THIS AND DO NOT ADD ANY OPTIONS TO IT
\DeclareCaptionStyle{ruled}{labelfont=normalfont,labelsep=colon,strut=off} % DO NOT CHANGE THIS
\usepackage{algorithm}
\usepackage{algorithmic}
\usepackage{pifont}
\usepackage{amsmath}
\usepackage{amsthm}
\usepackage{makecell}
\usepackage{multirow}
\usepackage{booktabs}       % professional-quality tables
\usepackage{subfigure}
\usepackage[title]{appendix}

\usepackage{url}            % simple URL typesetting
\usepackage{newfloat}
\usepackage{listings}
\floatstyle{ruled}
\newfloat{listing}{tb}{lst}{}
\floatname{listing}{Listing}
\title{FedLPS: Heterogeneous Federated Learning for Multiple Tasks\\with Local Parameter Sharing}
\author {
    Yongzhe~Jia\textsuperscript{\rm 1}, 
    Xuyun~Zhang\textsuperscript{\rm 2,*}, 
    Amin~Beheshti\textsuperscript{\rm 2}, 
    Wanchun~Dou\textsuperscript{\rm 1,}\thanks{Corresponding authors.}
}
\affiliations {
    \textsuperscript{\rm 1} State Key Laboratory for Novel Software Technology,\\
Department of Computer Science and Technology, Nanjing University, China\\
    \textsuperscript{\rm 2} School of Computing, Macquarie University, Australia\\
    jiayz@smail.nju.edu.cn, 
    \{xuyun.zhang, amin.beheshti\}@mq.edu.au, douwc@nju.edu.cn
}
\begin{document}

\maketitle

\begin{abstract}

Federated Learning (FL) has emerged as a promising solution in Edge Computing (EC) environments to process the proliferation of data generated by edge devices. By collaboratively optimizing the global machine learning models on distributed edge devices, FL circumvents the need for transmitting raw data and enhances user privacy. 
Despite practical successes, FL still confronts significant challenges including constrained edge device resources, multiple tasks deployment, and data heterogeneity.
% Firstly, the inherent limitations of edge devices, featured by constrained storage and computational resources, hinder the efficiency of training large-scale Deep Neural Networks (DNNs). Secondly, the prevalence of deploying multiple tasks on single devices has necessitated the maintenance of task-specific DNN models, exacerbating computational resource demands. Lastly, data heterogeneity across edge devices, coupled with variations in system capabilities, introduces complexities that impact model performance. 
However, existing studies focus on mitigating the FL training costs of each single task whereas neglecting the resource consumption across multiple tasks in heterogeneous FL scenarios. In this paper, we propose heterogeneous FEDerated learning with Local Parameter Sharing (FedLPS) to fill this gap. FedLPS leverages principles from transfer learning to facilitate the deployment of multiple tasks on a single device by dividing the local model into a shareable encoder and task-specific predictors. To further reduce resource consumption, a channel-wise model pruning algorithm that shrinks the footprint of local models while accounting for both data and system heterogeneity is employed in FedLPS. Additionally, a novel heterogeneous model aggregation algorithm is proposed to aggregate the heterogeneous predictors in FedLPS.
We implemented the proposed FedLPS on a real FL platform and compared it with state-of-the-art (SOTA) FL frameworks. The experimental results on five popular datasets and two modern DNN models illustrate that the proposed FedLPS significantly outperforms the SOTA FL frameworks by up to 4.88\% and reduces the computational resource consumption by 21.3\%. Our code is available at: \url{https://github.com/jyzgh/FedLPS}.
% \url{https://anonymous.4open.science/r/FedLPS}.

\end{abstract}

\section{Introduction}
% FL
Over the past decade, there has been a remarkable surge in the generation of massive amounts of data from billions of Internet of Things (IoT) devices~\cite{khan2021federated}. In the context of Edge Computing (EC) environments, Federated Learning (FL) has emerged as a promising solution for processing such extensive data at the edge~\cite{fedavg, khan2020federated}. FL enables various edge devices to collaboratively optimize a global Machine Learning (ML) model with the assistance of an edge server. This approach entails clients (i.e., edge devices) updating the ML model using their private local data, while the central server is responsible for aggregating these updated local models. In contrast to traditional centralized ML, FL facilitates data processing on distributed edge devices where the data is generated. In addition, the distributed learning nature of FL eliminates the need for transmitting raw data to the central server, thus avoiding unnecessary communication costs and enhancing user privacy~\cite{fedavg, ezzeldin2023fairfed}.

% Challenges
Despite the successful practical applications in FL, it still faces the following critical challenges:
\textbf{a) Limited resource budgets of edge devices.} Popular Deep Neural Networks (DNNs) commonly possess larger than hundreds of megabytes of parameters that need to be trained~\cite{resnet}, whereas the computational resources of edge devices are limited in EC environments~\cite{kairouz2021advances}. Training these unleashing DNN models on edge devices heavily hampered the learning efficiency of FL.
\textbf{b) Multitasking on a single device.} Ideally, each edge device maintains a task-specific DNN model for each task, resulting in linear growth of training costs on edge devices. Therefore, directly deploying multiple unleashing DNN models for each specific task on edge devices is impractical~\cite{fu2021l2s, ma2019snr, wallingford2022TAPS}.
\textbf{c) Data heterogeneity and system heterogeneity.} On one hand, the numbers of data samples and the data distributions of edge devices are typically various (i.e., non-identically distributed data, non-IID data), resulting in the accuracy degradation of the global model~\cite{gao2022survey, LDA}. On the other hand, the system capabilities of edge devices, such as CPU, GPU, memory, battery power, etc., also can be various. The devices with weaker system capabilities (i.e., ``stragglers'') will fail to finish local training, therefore hampering the learning process of FL~\cite{gao2022survey, fedmp}.

% Existing work and their limitations 
% [x,x,x] address challenge a), whereas neglecting xxx,
% [x,x,x] address challenge b), whereas neglecting xxx,
% % [x,x,x] address challenge c), whereas neglecting xxx,
% Moreover, few of them perform resource consumption reduction across multiple tasks on edge devices.
% Morden mobile devices in edge computing commonly run multiple tasks on a single device. In traditional ML, to realize satisfying ACC, xxx specific model. However, it is not suitable for edge devices since xxx. 

Several pioneering work have made efforts to mitigate these challenges through various solutions~\cite{FedDrop2018, FedProx2020, Wang2020Fednova, fedmp, li2021hermes}. 
% Caldas {et al.}~\cite{FedDrop2018} propose FedDrop to reduce the computational burden of local training and the corresponding communication costs of FL. FedDrop leverages lossy compression techniques to shrink the footprint of the ML model and generates identical compact local models for all devices. However, the identical models adopted by FedDrop fail to address either the data heterogeneity problem or the system heterogeneity problem, leading to a large accuracy loss on the global model. 
Li {et al.}~\cite{FedProx2020} propose FedProx to address the data heterogeneity problem by incentivizing participants to preserve similarity with the global model, while addressing the system heterogeneity problem by accommodating low-end devices (i.e., devices with fewer system capabilities) to carry out a reduced number of local updates. However, FedProx does not involve model compression techniques, making it difficult to save storage resources or communication overhead.
Recent work~\cite{fedmp, li2021hermes} propose to leverage model pruning techniques and personalized model aggregation to reduce the footprint of the ML model meanwhile mitigating the model accuracy loss caused by the heterogeneity problems.
% neglecting potential relations and similarities between the multiple ML models used for multiple tasks. 
However, these approaches focus only on each single ML model used for a specific task and thus contribute limitedly to reducing resource consumption across multiple tasks.

% How PS works in FL and its advantages
% Inspired by TL, we leverage PS to xxx. However, applying PS in FL is non-trivial due to edge devices' heterogeneous nature and their strict resource constraints. 

In this paper, we propose heterogeneous FEDerated learning with Local Parameter Sharing (FedLPS), a novel FL framework for reducing resource consumption of multiple tasks in the heterogeneous FL environment. Specifically, we leverage the spirit of transfer learning~\cite{Yosinski2014transferable, zhuang2020comprehensive} to allow the multiple tasks on a single device to share partial parameters of the ML models during the FL process. In contrast to existing FL frameworks, the proposed FedLPS reduces the resource consumption of devices not only in the context of a single task but also in the context of multiple tasks. We adopt channel-wise model pruning techniques to reduce the footprint of the local models and satisfy the strict resource budget of edge devices. Different from pioneering work that use uniform model pruning techniques (e.g., FedDrop~\cite{FedDrop2018}), FedLPS generates tailored models for each participant device to mitigate data heterogeneity problem and system heterogeneity problem. Moreover, considering that aggregating the tailored models with the popular FL aggregation algorithms (e.g., FedAvg~\cite{fedavg}) leads to the degradation of model performance, we further design a heterogeneous aggregation algorithm for FedLPS to generate the aggregated global model. We summarize our contributions as follows:

\begin{itemize}
\item We propose a novel FL framework FedLPS to reduce resource consumption of edge devices that deployed with multiple tasks. By dividing the local models on edge devices into shared encoders and task-specific predictors, FedLPS reduces the training cost across multiple tasks on edge devices.

\item We design a channel-wise model pruning algorithm for FedLPS to reduce the footprints of the predictors. By applying various pruning ratios, FedLPS adaptively shrinks model footprints of heterogeneous devices.

\item We present a heterogeneous model aggregation algorithm for FedLPS to aggregate heterogeneous task-specific predictors. By utilizing the knowledge within the pre-trained backbone model, FedLPS efficiently aggregates the heterogeneous predictors.
% By aggregating the parameters of local models in an ordered manner, FedLPS mitigates model accuracy loss caused by traditional element-wise weight averaging.

\item We implement the proposed FedLPS in a real-world FL platform FedML and extensively evaluate FedLPS with state-of-the-art FL frameworks. The experimental results demonstrated that FedLPS is effective in reducing resource consumption of edge devices while realizing heightened model accuracy.

\end{itemize}

\section{Related Work}

\subsection{Heterogeneous Federated Learning}

In the context of edge computing, federated learning is proposed to train ML models with distributed local data among edge devices~\cite{fedavg, khan2020federated, gao2022survey, zhang2023fedala}. In federated learning, the raw data on edge devices will be kept locally to provide better user privacy and avoid unnecessary communication costs, and only intermediate results (e.g., parameters of models) are transmitted between the server and the devices~\cite{khan2021federated}. Federated learning in the edge computing environments is commonly heterogeneous in several aspects such as statistics, systems, data spaces, and models~\cite{gao2022survey}. In this paper, we focus on two of the main heterogeneous aspects: statistical heterogeneity and system heterogeneity. Statistical heterogeneity leads to non-IID distribution of data among edge devices, while system heterogeneity leads to variations in the capabilities of these devices.
% With statistical heterogeneity, the local data can be non-IID distributed on edge devices, and system heterogeneity leads to variations in the capabilities of edge devices. 
% Both statistical heterogeneity and system heterogeneity impose significant challenges on federated learning.

\subsection{Transfer Learning}
% (In the inter-device context, TL has been investigated to address the xxx problem. However, little existing work has explored its feasibility for reducing resource consumption within devices running multiple tasks.)
Transfer learning is a promising machine learning methodology for transferring knowledge across different domains following different probability distributions~\cite{Yosinski2014transferable, long2017deep, zhuang2020comprehensive, tan2023federated}. Di {et al.}~\cite{di2017cross} propose to transfer the knowledge of images that are taken from a certain location, aiming to alleviate the adverse impact caused by various conditions such as different weather and illumination conditions in transportation applications. Yu {et al.}~\cite{yu2022spatl} propose SPATL for addressing the resource consumption problem and data heterogeneity problem in FL. In SPATL, transfer learning is adopted to address the data heterogeneity problem by transferring the knowledge of a shared encoder to the predictors on heterogeneous clients. Tu {et al.}~\cite{tu2021feddl} propose FedDL to capture the potential relationships between users and transfer knowledge between the related users in FL, aiming to improve the performance of Human Activity Recognition (HAR) task with unbalanced and sparse user data. 
However, few existing work explore the transferability of ML models for multiple tasks on edge devices in FL. In contrast, we fill this gap in this paper and demonstrate the feasibility of leveraging transfer learning to reduce resource consumption of edge devices deployed with multiple tasks.

\subsection{Model Pruning}

\begin{figure*}[!t]
\centerline{\includegraphics[width=1\textwidth]{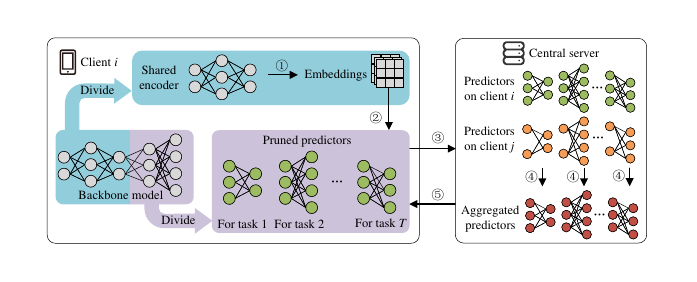}}
\caption{Overview of the proposed FedLPS framework. In FedLPS, the backbone model within each client is divided into the shared encoder and task-specific predictors. The predictors subsequently pruned to reduce resource consumption. During the training process, the encoder parameters remain frozen, while the predictor parameters are updated to handle specific tasks and transmitted between the central server and the client. To elaborate, \ding{172} local data for each task is fed into the encoder to generate embeddings. \ding{173} The task-specific predictors utilize these embeddings to update their parameters. \ding{174} The client sends the updated predictors to the central server. \ding{175} The central server aggregates the predictors that have been updated on different clients but belong to the same task. \ding{176} The central server sends the aggregated predictors back to the clients for further training rounds.}
\label{fig_overview}
\end{figure*}

Model pruning techniques are proposed to accelerate the training and inference processes of DNN by removing the redundant parameters and structures in the DNN model~\cite{liu2017slimming, he2017channel, li2021hermes, fedmp, ye2023performance}. Liu {et al.}~\cite{liu2017slimming} propose the network slimming scheme for Convolutional Neural Networks (CNNs), aiming to identify and remove insignificant parameters in CNNs by imposing L1 regularization on the scaling factors in batch normalization layers. Caldas {et al.}~\cite{FedDrop2018} propose FedDrop to reduce the computational burden of local training and the corresponding communication costs of FL. FedDrop leverages lossy compression techniques to shrink the footprint of the ML model and generates identical compact local models for all devices. Jiang {et al.}~\cite{fedmp} propose FedMP to address system heterogeneity problem meanwhile saving communication bandwidth. FedMP adopts a multi-armed bandit-based online learning algorithm to calculate personalized pruning ratios for heterogeneous edge devices and a Residual Recovery Synchronous Parallel (R2SP) scheme to aggregate parameters.
However, most of the existing work is on model pruning of native ML models on edge devices, and very little work has focused on transferable models applicable to multi-task scenarios. In contrast, we design an adaptive channel-wise model pruning algorithm for the transferable models in FedLPS to reduce unnecessary resource consumption in multi-task scenarios.

\section{Design of FedLPS}

In this section, we first present an overview of the proposed FedLPS. Then, we describe how transfer learning can be used to train models for multiple tasks. Subsequently, we employ an adaptive channel-wise model pruning approach to reduce the resource consumption caused by training task-specific predictors. Finally, we present a heterogeneous model aggregation algorithm for aggregating heterogeneous predictors updated by different clients.

\subsection{Overview}

In this paper, we propose FedLPS to efficiently train multiple task-specific models on individual clients in the context of FL. Fig.~\ref{fig_overview} illustrates the workflow of the proposed FedLPS framework. Distinct from existing FL frameworks that optimize a specialized model for each task on the client, in FedLPS, the backbone model $\boldsymbol{w}^i$ in individual client $i \in \mathcal{C}$ is partitioned into a shared encoder $\boldsymbol{w}_e^i$, and $|\mathcal{T}|$ task-specific predictors $\boldsymbol{w}_p^{i,t}$ for each task $t \in \mathcal{T}$, where $\mathcal{T}$ denote the task set. The backbone model can be either trained on the public dataset or trained on local data of arbitrary tasks on the client and is accessible to both the client and the central server. During the federated training process, the encoder $\boldsymbol{w}_e^i$ will be frozen while the task-specific predictor $\boldsymbol{w}_p^{i,t}$ will be updated on the local data of the task $t$. 

We outline the federated training steps as follows: \ding{172} The local data for each task is fed into the encoder $\boldsymbol{w}_e^i$ to generate embeddings. \ding{173} The task-specific predictors $\boldsymbol{w}_p^{i,t}$ utilize these embeddings generated by local data of tasks $t$ to update to update their parameters. \ding{174} Each client $i$ sends the updated predictors $\boldsymbol{w}_p^{i,t}$ to the central server. \ding{175} The central server aggregates the predictors that have been updated on different clients (e.g., predictors on client $i$ and predictors on client $j$ in Fig.~\ref{fig_overview}) but belong to the same task $t$. \ding{176} The central server sends the aggregated predictors back to the clients for further training rounds.

\subsection{Local Parameter Sharing across Multiple Tasks}

Existing FL frameworks (e.g., FedAvg~\cite{fedavg}, FedDrop~\cite{FedDrop2018}, Hermes~\cite{li2021hermes}) commonly adopt an approach where a specific model is optimized for each FL task on the client. However, this practice of optimizing multiple models on the clients leads to significant resource consumption, especially for edge devices with limited capabilities. Although some efforts~\cite{FedProx2020, fedmp} have been made to reduce the training overhead on the client, few of them focus on effectively reducing training overhead across multiple FL tasks. In this subsection, we propose a novel federated training method that leverages local parameter sharing on the clients to mitigate the resource consumption associated with training multiple models for multiple tasks, meanwhile maintaining a satisfying model accuracy.
\footnote{In this work, we make the fundamental assumption that the model structure remains consistent across all tasks, and this work does not focus on addressing the issue of model heterogeneity.}

Inspired by the spirit of the transfer learning~\cite{zhuang2020comprehensive, weiss2016survey, Yosinski2014transferable}, we explore enabling the multiple tasks on each client to share a part of model parameters, thus effectively reducing the training overhead. In parameter-sharing-based transfer learning, the lower layers of the neural networks capture more generalized features, making them suitable for sharing across multiple tasks. Conversely, the upper layers tend to capture higher-level abstract features, making them more task-specific. Building upon this observation, we divide the local backbone model $\boldsymbol{w}$ on each client into a shareable encoder $\boldsymbol{w}_e$ and multiple task-specific predictors $\boldsymbol{w}_p^{t}$. 
\footnote{For presentation convenience, we omit the superscript $i$ when describing local operation of each client.}
The shared encoder $\boldsymbol{w}_e$ comprises the first $n$ layers of the backbone model, while the task-specific predictors $\boldsymbol{w}_p^{t}$ consist of the remaining $N-n$ layers, where $n$ is a tunable hyper-parameter and the $N$ represents the total number of layers in the backbone model. In order to facilitate knowledge transfer across various tasks, the weights of shared encoder $\boldsymbol{w}_e$ are initialized using pre-trained values. Throughout the training process, the weights in shared encoder $\boldsymbol{w}_e$ remains frozen, ensuring consistent utilization across all tasks. The pre-trained weights can be sourced either from a backbone model trained on publicly available datasets or from client-participated FL tasks. In addition, to reduce the resource consumption caused by training multiple predictors, an adaptive channel-wise model pruning method is proposed for shrinking the footprints of the predictors, which is described in the next subsection in detail.

The local training algorithm based on local parameter sharing of the proposed FedLPS framework is presented in Algorithm~\ref{alg_train}. In each communication round of FL, the client first prunes each task-specific predictor $\boldsymbol{w}_p^{t}$ with a pruning ratio $\rho$, aiming to reduce subsequent training costs (in line 2). Then, the client conducts forward propagation on the shared encoder $\boldsymbol{w}_e$ using local data $\mathcal{D}_t$ specific to each task $k$. The resulting embedding is denoted as $\boldsymbol{e}^t$ (in line 3). Subsequently, each pruned predictor ${\boldsymbol{w}_p^{t}}'$ is updated with the embedding $\boldsymbol{e}^t$ (in line 4). Specifically, the update operation can be formulated as follows:
\begin{equation}
{\boldsymbol{w}_p^{t}}' \leftarrow {\boldsymbol{w}_p^{t}}' - \eta \nabla \ell_t({\boldsymbol{w}_p^{t}}'), \label{eq_sgd}
\end{equation}
where $\eta$ is the learning rate, $\ell_t$ is the loss function of task $t$, and $\nabla \ell_t({\boldsymbol{w}_p^{t}}')$ is the local gradients of the predictors ${\boldsymbol{w}_p^{t}}'$. 

Finally, the updated task-specific predictors ${\boldsymbol{w}_p^{t}}'$ will be sent to the central server for aggregation.

\begin{algorithm}[t]
	\caption{Local Parameter Sharing-based Training Algorithm of FedLPS.}\label{alg_train}
	\begin{algorithmic}[1]
	\renewcommand{\algorithmicrequire}{\textbf{Input:}}
	\renewcommand{\algorithmicensure}{\textbf{Output:}}
	\REQUIRE Task set $\mathcal{T}$, local data $\mathcal{D}_t (t \in \mathcal{T})$, pre-trained encoder $\boldsymbol{w}_e$, pruning ratio $\rho$ for pruning predictors
	\ENSURE Updated predictors ${\boldsymbol{w}_p^{t}}'$
            % --------------------------Device Updates--------------------------
            % \\\textbf{Client Updates:}
            \FOR {task $t \in \mathcal{T}$}     
                \STATE ${\boldsymbol{w}_p^{t}}' \leftarrow$ Prune the predictor $\boldsymbol{w}_p^{t}$ with pruning ratio $\rho$ using Eq.~\ref{eq_pruning}
        	\STATE $\boldsymbol{e}^t \leftarrow$ Forward propagation on shared encoder $\boldsymbol{w}_e$ with local data $\mathcal{D}_t$
                % \STATE $\hat{\boldsymbol{y}}_t \leftarrow$ Forward propagation on predictor $\boldsymbol{w}_p^{t}$ with embedding $\boldsymbol{e}^t$                
                \STATE ${\boldsymbol{w}_p^{t}}' \leftarrow$ Update weights in predictor ${\boldsymbol{w}_p^{t}}'$ using Eq.~\ref{eq_sgd}
            \ENDFOR
            \RETURN Updated predictors ${\boldsymbol{w}_p^{t}}'$      
	\end{algorithmic}
\end{algorithm}

\subsection{Adaptive Channel-wise Model Pruning}

By leveraging the division of the backbone model and the freezing of the pre-trained shared encoder, FedLPS effectively mitigates the training cost associated with the shared encoder. However, the training costs incurred by the predictors continue to pose challenges for clients due to the generally larger footprints associated with these predictors compared to the encoder. Thus, in this subsection, we proposed an adaptive channel-wise model pruning method for FedLPS to reduce the training cost of the task-specific predictors.

In the first communication round of FL, each client prunes the predictors with a punning ratio $\rho$. Differing from existing model pruning methods in FL that adopt a uniform punning ratio for each client like FedDrop~\cite{FedDrop2018}, FedLPS prunes the predictors on heterogeneous clients with different punning ratios $\rho$ that are determined by the clients' system capability. Firstly, for each task $t \in \mathcal{T}$ on the client, FedLPS evaluates the importance scores of channels in each layer of the predictor $\boldsymbol{w}_p^{t}$ by L1-norm~\cite{liu2017slimming, li2016pruning}. Then, a fraction $\rho$ of channels corresponding to the smallest importance scores is removed to achieve model pruning. Specifically, the pruning operation can be formulated as follows:
\begin{equation}
{\boldsymbol{w}_p^{t}}' \leftarrow \boldsymbol{w}_p^{t} \odot \boldsymbol{M^t}, \boldsymbol{M^t} \in \{ 0,1 \}^{\lvert \boldsymbol{w}_p^{t} \rvert},      \label{eq_pruning}
\end{equation}
where $\odot$ denotes the element-wise multiplication, and $\boldsymbol{M^t}$ is a binary mask matrix used to determine the channels to be pruned. In the mask matrix $\boldsymbol{M^t}$, elements with a value of 0 indicate channels that will be pruned, while elements with a value of 1 indicate channels that will be retained. Finally, the pruned predictors ${\boldsymbol{w}_p^{t}}'$ will be updated with Eq.~\ref{eq_sgd}.

\begin{algorithm}[t]
	\caption{Heterogeneous Predictor Aggregation Algorithm of FedLPS.}\label{alg_agg}
	\begin{algorithmic}[1]
	\renewcommand{\algorithmicrequire}{\textbf{Input:}}
	\renewcommand{\algorithmicensure}{\textbf{Output:}}
	\REQUIRE Pruned predictors ${\boldsymbol{w}_p^{i,t}}' (i \in \mathcal{C}, t \in \mathcal{T})$, mask matrices $\boldsymbol{M}^{i,t}$, backbone predictor $\boldsymbol{w}_p$
	\ENSURE Aggregated predictors ${\boldsymbol{w}_p^{t}}^*$
            % \\\textbf{Server Excutes:}
            \FOR {task $t \in \mathcal{T}$}  
                \FOR {client $i \in \mathcal{C}$} 
                    \STATE ${\boldsymbol{w}_p^{i,t}} \leftarrow$ Recover pruned predictor ${\boldsymbol{w}_p^{i,t}}'$ using backbone predictor $\boldsymbol{w}_p$ and mask matrix $\boldsymbol{M}^{i,t}$ (Eq.~\ref{eq_recover})
                \ENDFOR    
                \STATE ${\boldsymbol{w}_p^{t}}^* \leftarrow$ Aggregate the recovered predictors of task $t$ using Eq.~\ref{eq_agg}
                % \STATE ${\boldsymbol{w}_p^{t}}' \leftarrow$ Update weights in predictor ${\boldsymbol{w}_p^{t}}'$ using Eq.~\ref{eq_sgd}
            \ENDFOR
            \RETURN Aggregated predictors ${\boldsymbol{w}_p^{t}}^*$    
	\end{algorithmic}
\end{algorithm}

\subsection{Heterogeneous Predictor Aggregation}

The adaptive pruning operation on the predictors is capable to shrink their footprint and thereby reducing the resource consumption of the heterogeneous clients. However, the adaptive pruning operation leads to multiple heterogeneous predictors that can not be aggregated by popular FL aggregation algorithms (e.g., FedAvg~\cite{fedavg}). Existing heterogeneous aggregation algorithms such as Hermes~\cite{li2021hermes} and FedMP~\cite{fedmp}, aggregate the overlapped parameters of the heterogeneous models. Unfortunately, these algorithms result in a degradation in model performance due to the pruned parameters lacking any significant contribution to the aggregated global model. Moreover, the remained parameters are difficult to learn knowledge from other clients when these parameters are pruned in other clients. Thus, in this subsection, we propose a novel aggregation algorithm that leverages the knowledge within the pre-trained backbone model to aggregate the heterogeneous predictors. 

The heterogeneous predictor aggregation algorithm of the proposed FedLPS framework is presented in Algorithm~\ref{alg_agg}. In each FL communication round, the central server receives $|\mathcal{T}|$ task-specific predictors from each selected client. FedLPS first recovers the pruned parameters in each task-specific predictor ${\boldsymbol{w}_p^{i,t}}'$ using the backbone predictor $\boldsymbol{w}_p$ extracted from the backbone model and the mask matrix $\boldsymbol{M}^{i,t}$ on task $t$ of client $i$ (in line 3). The recovery operation can be formulated as follows:
\begin{equation} \label{eq_recover}
{\boldsymbol{w}_p^{i,t}} \leftarrow {\boldsymbol{w}_p^{i,t}}' + [\boldsymbol{w}_p - (\boldsymbol{w}_p \odot \boldsymbol{M}^{i,t})],      
\end{equation}
where $\boldsymbol{w}_p^{i,t}$ represents the recovered predictor containing the updated weights on task $t$ of client $i$. 
\footnote{It is noteworthy that these updated weights differ from the initial predictor, despite using the same symbol.}
Subsequently, FedLPS can aggregate these recovered predictors of task $t$ by weighted averaging (in line 5). Formally, the aggregation operation can be represented as follows:
\begin{equation}    \label{eq_agg}
{\boldsymbol{w}_p^{t}}^* \leftarrow \sum_{i \in \mathcal{C}} \frac{\lvert \mathcal{D}_{i,t} \rvert}{\lvert \mathcal{D}_{t} \rvert} {\boldsymbol{w}_p^{i,t}},
\end{equation}
where $\mathcal{D}_{t}$ denotes the entire data of task $t$ across all clients and $\mathcal{D}_{i,t}$ denotes the local data of task $t$ on the client $i$. For each task $t \in \mathcal{T}$, FedLPS aggregates the predictor ${\boldsymbol{w}_p^{i,t}}$ using Eq.~\ref{eq_agg} until all predictors have been aggregated.

Finally, the central server sends the aggregated global predictors ${\boldsymbol{w}_p^{i,t}}$ back to the selected clients for the further round of local training.

\section{Experimental Evaluation}

In this section, we first implement the proposed FedLPS framework on a real federated learning platform FedML~\cite{fedml} and conduct a comprehensive performance comparison of FedLPS against five state-of-the-art (SOTA) frameworks in multiple tasks-enabled FL environments. 
% Subsequently, we conduct an in-depth analysis of the resource consumption of the FedLPS framework, highlighting its distinct advantages in reducing training costs. 
Next, we evaluate the effect of varying pruning ratios on the learning performance of the FedLPS. Finally, we evaluate the effect of the layer number $n$ of the shared encoder on the learning performance of the FedLPS.

\subsection{Experimental Setting}

\subsubsection{FL environments.}
% We build an FL environment with the IID data setting and the other with the non-IID data setting on FedML~\cite{fedml} to evaluate the proposed FedLPS framework. The details of both environments are provided in Table~\ref{tab_setting}. 
In our experiments, we simulate 10 heterogeneous clients and deploy 5 classification tasks on each client. Both the ResNet18 model~\cite{resnet} and the ShuffleNetV2 model~\cite{zhang2018shufflenet} are adopted to conduct these classification tasks. The heterogeneous client set $\mathcal{C}$ are uniformly divided into 5 levels according to their system capabilities, denoted as $\mathcal{C} = \{\mathcal{C}_1,\mathcal{C}_2,\mathcal{C}_3,\mathcal{C}_4,\mathcal{C}_5\}$. The system capabilities are decreasing from $\mathcal{C}_1$ to $\mathcal{C}_5$ linearly.
% , denoted as $\{S, 0.8S, 0.6S, 0.4S, 0.2S\}$. 
Only the clients in $\mathcal{C}_1$ can conduct these 5 classification tasks without resource optimizing techniques (i.e., model pruning).
% Clients in $\mathcal{C}_1$ have the most powerful system capabilities while the clients in $\mathcal{C}_5$ have the least system capabilities. 
All the heterogeneous clients are selected to perform FL training. 
The experiments are conducted on a GPU server with 2 NVIDIA RTX 3080Ti GPUs, and each experiment is executed three times for calculating average metrics.
% where ResNet18 is a popular deep learning model containing 11.18M parameters and ShuffleNetV2 is a lightweight deep learning model containing 1.26M parameters.

% \begin{table}[ht]
% \caption{Data partition, models, client number, and task number of the FL environments used in our experiments. ENV is the abbreviation of environment, NIID is the abbreviation of non-IID, and num. is the abbreviation of number.}\label{tab_setting}
% \centering\begin{tabular}{lcccc}
% \toprule
% FL ENVs   & \makecell{Data\\partition} & Models                               & \makecell{Task\\num.} & \makecell{Client\\num.}     \\
% \midrule                                                                                                                                  
% ENV-IID   & IID                        & \makecell{ResNet18 \&\\ShuffleNetV2} &           5           &         10                  \\
% ENV-NIID  & NIID                       & \makecell{ResNet18 \&\\ShuffleNetV2} &           5           &         10                  \\
% \bottomrule
% \end{tabular}
% \end{table}

\begin{table*}[!t]
% MNIST, FashionMNIST, SVHN, CIFAR10, and CIFAR100 datasets.}\label{tab_test_acc}
\centering\begin{tabular}{llcccccc}
\toprule
{\makecell{Data\\partition}} & {FL frameworks}  
                               & {MNIST} & {\makecell{Fashion-\\MNIST}} & {SVHN} & {CIFAR10} & {CIFAR100}  & {Average} \\
\midrule
\multirow{6}{*}{\makecell{IID\\partition}}
&FedAvg\cite{fedavg}           &  97.07  &         86.35  &         89.32  &         64.15  &         25.27  & 72.43 \\
&FedDrop\cite{FedDrop2018}     &  77.92  &         73.26  &         37.28  &         56.77  &         23.97  & 53.84 \\
&FedProx\cite{FedProx2020}     &  96.63  &         86.10  &         89.41  &         73.88  &         38.65  & 76.93 \\
&Hermes\cite{li2021hermes}     &  97.92  &         87.73  &         91.24  &         76.32  &         38.92  & 78.43 \\
&FedMP\cite{fedmp}             &  97.08  &         87.29  &         88.45  &         74.20  &         38.78  & 77.16 \\\cmidrule(lr){2-8}
&\textbf{FedLPS (Ours)} & \textbf{97.99} & \textbf{87.94} & \textbf{91.39} & \textbf{78.71} & \textbf{49.66} & \textbf{81.14} \\
\midrule
\multirow{6}{*}{\makecell{Non-IID\\partition}}
&FedAvg\cite{fedavg}           &  56.03  &          59.36  &         79.23  &         26.39  &         20.04 & 48.21 \\
&FedDrop\cite{FedDrop2018}     &  77.55  &          61.81  &         36.37  &         41.91  &         23.06 & 48.14 \\
&FedProx\cite{FedProx2020}     &  93.77  &          83.16  &         82.96  &         62.14  &         37.17 & 71.84 \\
&Hermes\cite{li2021hermes}     &  95.32  &          82.53  &         86.29  &         60.09  &         38.26 & 72.50 \\
&FedMP\cite{fedmp}             &  95.60  &  \textbf{84.21} &         82.34  &         61.59  &         37.19 & 72.19 \\\cmidrule(lr){2-8}
&\textbf{FedLPS (Ours)} & \textbf{96.67} &          83.39  & \textbf{88.00} & \textbf{71.15} & \textbf{47.70} & \textbf{77.38} \\
\bottomrule
\end{tabular}
\caption{Comparison of model accuracy (\%) on the ShuffleNetV2 model with both IID and non-IID data.}\label{tab_sfnet_acc}
\end{table*}

\begin{table*}[!t]
% MNIST, FashionMNIST, SVHN, CIFAR10, and CIFAR100 datasets.}\label{tab_test_acc}
\centering\begin{tabular}{llcccccc}
\toprule
{\makecell{Data\\partition}} & {FL frameworks}  
                               & {MNIST} & {\makecell{Fashion-\\MNIST}} & {SVHN} & {CIFAR10} & {CIFAR100}  & {Average} \\
\midrule
\multirow{6}{*}{\makecell{IID\\partition}}
&FedAvg\cite{fedavg}           &  98.38  &         88.82  &         93.74  &         76.74  &         35.06  & 78.55  \\
&FedDrop\cite{FedDrop2018}     &  88.59  &         83.19  &         69.37  &         58.39  &         23.41  & 64.59  \\
&FedProx\cite{FedProx2020}     &  98.60  &         90.30  &         95.05  &         85.31  &         54.37  & 84.73  \\
&Hermes\cite{li2021hermes}     &  98.67  &         91.32  &         95.46  &         81.58  &         54.54  & 84.31  \\
&FedMP\cite{fedmp}             &  97.67  &         89.25  &         91.25  &         79.29  &         46.10  & 80.71  \\\cmidrule(lr){2-8}
&\textbf{FedLPS (Ours)} & \textbf{98.85} & \textbf{92.15} & \textbf{95.48} & \textbf{86.98} & \textbf{57.87} & \textbf{86.27} \\
\midrule
\multirow{6}{*}{\makecell{Non-IID\\partition}}
&FedAvg\cite{fedavg}           &  65.82  &         66.38  &         84.58  &         31.52  &         27.59  & 55.18  \\
&FedDrop\cite{FedDrop2018}     &  82.19  &         77.24  &         69.05  &         42.55  &         19.73  & 58.15  \\
&FedProx\cite{FedProx2020}     &  95.87  &         84.07  &         91.05  &         67.60  &         53.18  & 78.35  \\
&Hermes\cite{li2021hermes}     &  97.05  &         84.00  & \textbf{92.12} &         71.81  &         53.53  & 79.70  \\
&FedMP\cite{fedmp}             &  96.60  &         83.78  &         87.66  &         67.42  &         42.53  & 75.60  \\\cmidrule(lr){2-8}
&\textbf{FedLPS (Ours)} & \textbf{96.98} & \textbf{86.45} &         91.98  & \textbf{74.17} & \textbf{55.56} & \textbf{81.08} \\
\bottomrule
\end{tabular}
\caption{Comparison of model accuracy (\%) on the ResNet18 model with both IID and non-IID data.}\label{tab_r18_acc}
\end{table*}

\subsubsection{Datasets and data partition.}
In our experiments, we adopt five widely recognized datasets: MNIST~\cite{MNIST}, FashionMNIST~\cite{xiao2017fashion}, SVHN~\cite{netzer2011reading}, CIFAR10~\cite{cifar}, and CIFAR100~\cite{cifar} datasets to simulate the classification tasks. In the IID setting, each dataset is equally assigned to the clients. In the non-IID setting, we use the Latent Dirichlet Allocation (LDA)~\cite{LDA, FedMA} method to build the non-IID data. In LDA, a concentration parameter $\alpha$ is used to control the data heterogeneity. We adopt the conventional setting of $\alpha = 0.5$ in our experiments to construct the non-IID data.

\subsubsection{Comparison frameworks.}
We compare the proposed FedLPS framework with FedAvg~\cite{fedavg}, FedDrop~\cite{FedDrop2018}, FedProx~\cite{FedProx2020}, Hermes~\cite{li2021hermes}, and FedMP~\cite{fedmp}. FedAvg is the classical FL framework that needs each client to update the entire model, thus can involve only the clients in $\mathcal{C}_1$ to perform FL training in the heterogeneous FL environment. FedDrop leverages model pruning techniques to generate a compact global model that can be updated by the clients in $\mathcal{C}_5$, thus involving all clients to participate in the FL training. FedProx encourages local models to maintain similarity with the global model by introducing a regularization term meanwhile allowing the clients in $\{\mathcal{C}_2,\mathcal{C}_3,\mathcal{C}_4,\mathcal{C}_5\}$ to perform fewer local updates. Hermes extracts a tailored sub-model for each client by structured pruning and aggregates only the intersection of the local models. FedMP prunes each local model with a dynamic pruning ratio in each round thereby enabling all clients to participate in the FL training.

To conduct fair comparisons, we adopt the same training hyper-parameters for FedLPS and the comparison frameworks in our experiments. The training hyper-parameters are provided in the supplemental materials in detail.

\subsection{Learning Performance across Multiple Tasks}

We compare the model accuracy of the FedLPS with FedAvg, FedDrop, FedProx, Hermes, and FedMP on both IID and non-IID data. In this experiment, the layer number of the shared encoder in FedLPS is set as $n = \lceil 0.25N \rceil$, i.e., the first $\lceil 0.25N \rceil$ layers within the backbone model are used to build the shared encoder. For FedLPS, Hermes, and FedMP frameworks, the pruning ratios $\rho$ adopted by the 5 levels of clients are $\{0, 0.2, 0.4, 0.6, 0.8\}$, respectively. The backbone model is pre-trained on ImageNet~\cite{ImageNet}. 

Table~\ref{tab_sfnet_acc} shows the accuracy on the ShuffleNetV2 model with both IID and non-IID data, while Table~\ref{tab_r18_acc} shows the accuracy on the ResNet18 model.
Additionally, we provide a comparative analysis of communication overhead in the supplemental materials, assessed through the footprints of the transmitted models.
On both the ShuffleNetV2 and ResNet18 models, FedLPS outperforms the comparison frameworks in terms of average model accuracy. The superiority of FedLPS can be attributed to three reasons: firstly, the pre-trained shareable encoder is sophisticated to extract the low-level features and produce general embedding from the local data. Secondly, the task-specific predictors are pruned elaborately to satisfy the resource constraints, and the pruned predictors are trained on each task separately therefore suitable for performing the specific tasks. Thirdly, the heterogeneous predictor aggregation algorithm used in FedLPS leverages the knowledge in the backbone model to assist the aggregation of the local predictors, thus making the predictors learn from other clients better.

\subsection{Effect of Pruning Ratios}

\begin{figure}[!t]
\centering
\subfigure[CIFAR10]{\includegraphics[width=0.23\textwidth]{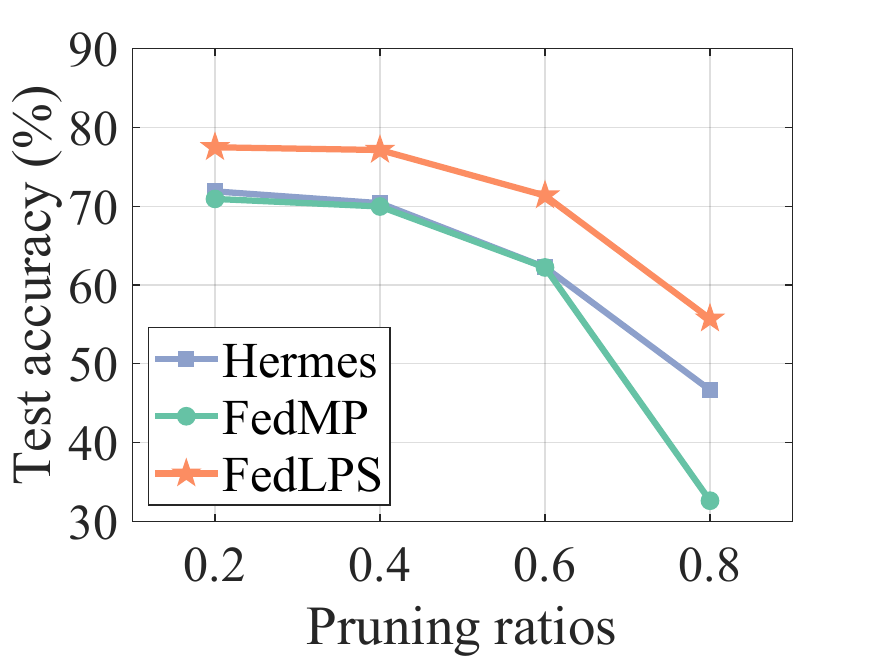}\label{fig_exp_pr_cf10}
}\subfigure[CIFAR100]{\includegraphics[width=0.23\textwidth]{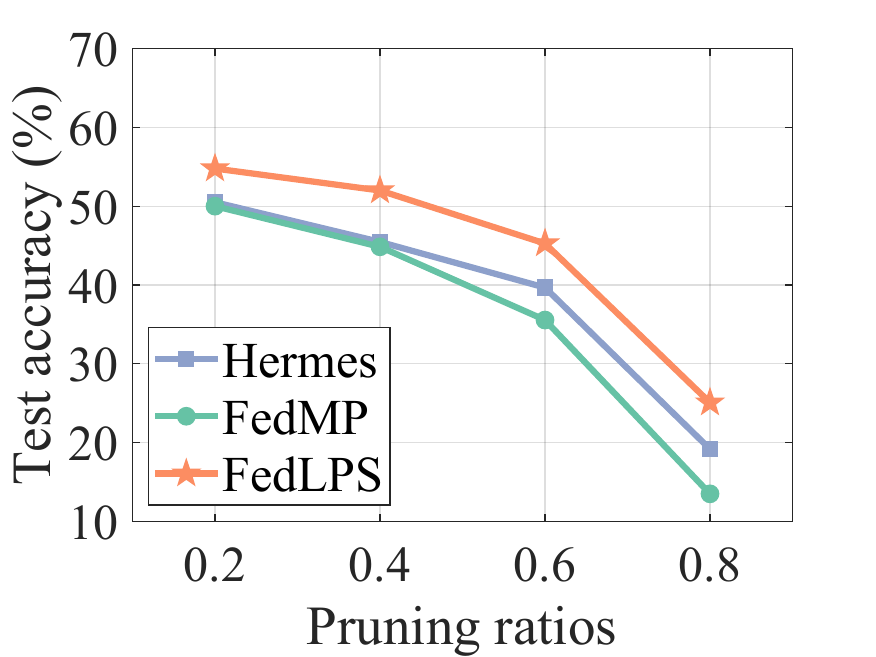}\label{fig_exp_pr_cf100}
}
\caption{Model accuracy of Hermes, FedMP, and FedLPS on the RseNet18 model with different pruning ratios $\rho$ on the non-IID setting of the CIFAR10 and CIFAR100 datasets.}
\label{fig_exp_pr}
\end{figure}

In the model pruning-enabled FL frameworks, the resource consumption of the training local model is significantly reduced, whereas the model accuracy will decrease when the pruning ratio exceeds a threshold. Thus, we investigate the effect of the pruning ratio on model accuracy in this subsection. In this experiment, the proposed FedLPS framework is compared with two SOTA model pruning-enabled frameworks Hermes and FedMP, and the pruning ratio $\rho$ is set as 0.2, 0.4, 0.6, and 0.8. Fig.~\ref{fig_exp_pr_cf10} and Fig.~\ref{fig_exp_pr_cf100} show the model accuracy of the ResNet18 model on the non-IID setting of the CIFAR10 dataset and the CIFAR100 dataset during 100 FL rounds, respectively. FedLPS outperforms Hermes and FedMP when the pruning ratio $\rho$ ranges from 0.2 to 0.8 in both the CIFAR10 dataset and the CIFAR100 dataset, although all of them adopt model pruning techniques to shrink the model footprint. The detailed footprints of the predictors within FedLPS are provided in the supplemental materials.
% Moreover, compared with the model accuracy obtained by training the entire model on the clients, FedLPS still achieves better performance when the pruning ratio $\rho$ is less than 0.4 on the CIFAR10 dataset and when $\rho=0.2$ on the CIFAR100 dataset.

\subsection{Effect of Local Parameter Sharing}

\begin{figure}[!t]
\centering
\subfigure[IID setting]{\includegraphics[width=0.23\textwidth]{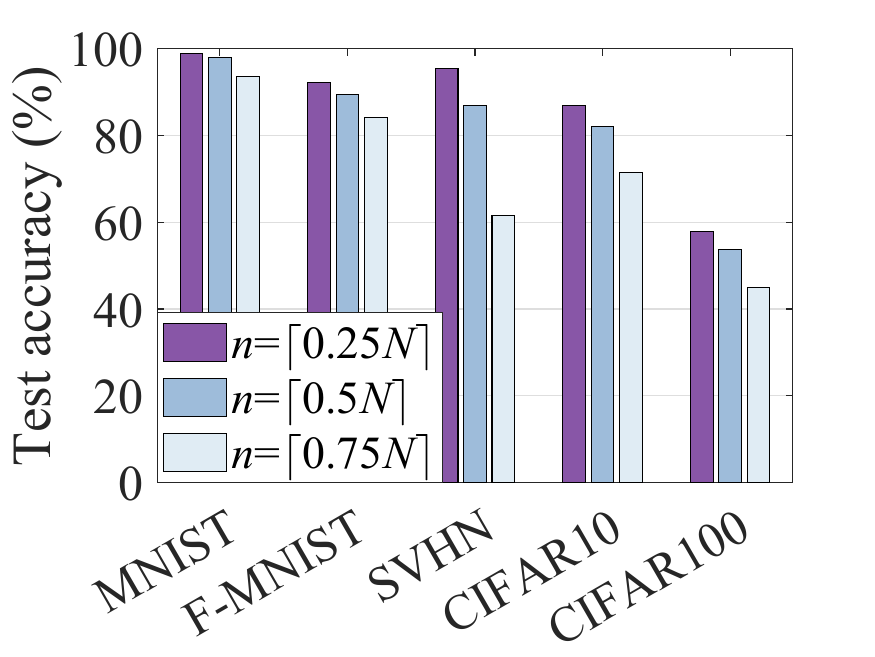}\label{fig_exp_frozen_iid}
}\subfigure[Non-IID setting]{\includegraphics[width=0.23\textwidth]{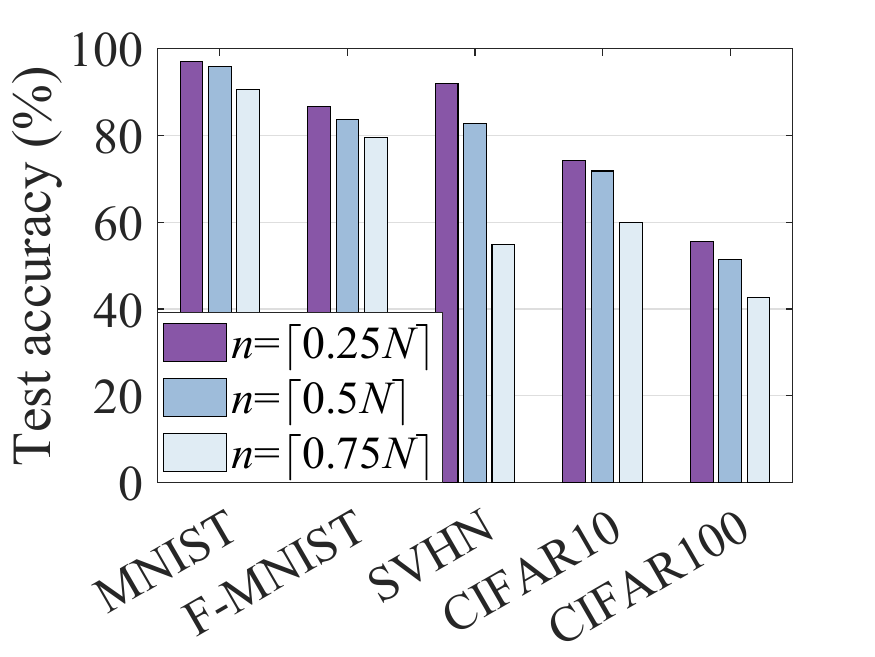}\label{fig_exp_frozen_niid}
}
\caption{Model accuracy of FedLPS with different layer numbers $n$ on the IID setting and the non-IID setting. Where F-MNIST represents FashionMNIST dataset.}
\label{fig_exp_frozen}
\end{figure}

In FedLPS, the larger the layer number $n$ assigned to the shared encoder, the more training costs can be saved. In this experiment, we investigate the effect of the layer number $n$ of the shared encoder on the training cost and model accuracy. The layer number $n$ is set as $\lceil 0.25N \rceil$, $\lceil 0.5N \rceil$, and $\lceil 0.75N \rceil$, respectively. The floating-point operations (FLOPs) of the model are measured to indicate the training cost. In this experiment, the ResNet18 model containing 34.36M FLOPs is adopted. 
Fig.~\ref{fig_exp_frozen} shows the model accuracy of MNIST, FashionMNIST, SVHN, CIFAR10, and CIFAR100 datasets while Table~\ref{tab_FLOPs} reports the FLOPs of the sum of the shared encoder and all task-specific predictors (before adaptive pruning). In FedLPS, the FLOPs linearly decrease as the layer number $n$ increases, resulting in a substantial reduction in training costs. 
Thus, FedLPS are more suitable for clients that are deployed with multiple tasks than existing FL frameworks as the training cost of multiple tasks can be easily reduced by building a shared encoder with $n$-layer. 
Take the clients in $\mathcal{C}_1$ that deployed with 5 tasks as an example, in Fig.~\ref{fig_exp_frozen} and Table~\ref{tab_FLOPs}, FedLPS achieves 86.27\% and 81.08\% of average accuracy with 21.3\% fewer FLOPs (i.e., $n=\lceil 0.25N \rceil$) in the IID setting and non-IID setting, respectively. Whereas the SOTA frameworks only realize 84.31\% and 79.70\% of average accuracy (in Table~\ref{tab_r18_acc}) with 171.8M FLOPs ($5 \times$ FLOPs of the original ResNet18 model). 

\begin{table}[!t]
% MNIST, FashionMNIST, SVHN, CIFAR10, and CIFAR100 datasets.}\label{tab_test_acc}
\centering\begin{tabular}{lccc}
\toprule
{\makecell{Layer\\number $n$}} & {FLOPs}  & {\makecell{Reduction\\on FLOPs}} & {\makecell{Average\\accuracy(\%)}}\\
\midrule
% Original ResNet18          &   171.8M  &      0  &  - \\
$n = \lceil 0.25N \rceil$  &   135.2M  &  -21.3\%  &  86.27(81.08)  \\
$n = \lceil 0.5N \rceil$   &   101.5M  &  -40.9\%  &  82.04(77.20)  \\
$n = \lceil 0.75N \rceil$  &    67.9M  &  -60.5\%  &  71.12(65.57)  \\
\bottomrule
\end{tabular}
\caption{Total FLOPs of FedLPS with different layer numbers $n$ and corresponding model average accuracy on both IID data and non-IID data. (The accuracy of IID data is outside the parentheses and the accuracy of non-IID data is inside the parentheses.)}\label{tab_FLOPs}
\end{table}

% \subsection{Resource Consumption on Clients}

% \subsection{Effect of Data Heterogeneity}

% \subsection{Effect of Client Selection Rate} % appendix

\section{Conclusion}
In this paper, we have proposed FedLPS for multiple-tasks-enabled heterogeneous FL environments, aiming to reduce the resource consumption of the clients during the FL training process while maintaining satisfying model accuracy. FedLPS realizes local parameter sharing by dividing the local model into a shareable encoder and multiple task-specific predictors, thus achieving the reduction of training costs across multiple tasks on individual clients. To tackle the system heterogeneity problem, an adaptive channel-wise model pruning method is proposed for FedLPS to allow the heterogeneous clients to participate in the FL training with heterogeneous task-specific predictors. Furthermore, a novel aggregation algorithm is proposed for FedLPS to efficiently aggregate the heterogeneous predictors with the assistance of the knowledge within the pre-trained backbone model. The comparison results on five popular datasets and two modern DNN models demonstrated the superiority of the FedLPS in terms of both average model accuracy and resource consumption.

\textbf{Limitations and Prospects for Future Research.} The hyper-parameter $n$ employed in FedLPS currently lacks the flexibility for dynamic adjustments during FL training, a promising future direction involves delving into the nuanced interactions among the layer number $n$, the pruning ratio $\rho$, and the ensuing model accuracy. Moreover, the scope of this study is confined to utilizing FedLPS exclusively for classification tasks. To extend the applicability of FedLPS, an exciting direction involves comprehensively exploring its performance across diverse task domains.

\section{Acknowledgments}
This research is supported part by the National Key Research and Development Program of China No.2020YFB1707601, National Natural Science Foundation of China No.92267104.
% This work is supported by the National Key Research and Development Project under Grant No. 2020YFB1707600.
% This research is supported part by Open Research Fund from Guangdong Laboratory of Artificial Intelligence and Digital Economy(SZ) No.GML-KF-22-20 and the Dou Wanchun Expert Workstation of Yunnan Province No.202105AF150013.

% \bigskip
% \balance
\bibliography{AAAI2024}

\clearpage
% \appendix
\begin{appendices}

% =================================== appendices ===================================

\section{Hyper-parameters used in FL training}\label{apd_hype}
% hyper-parameters

We elaborate on the training hyper-parameters employed in the experiments. Both the SOTA frameworks and our proposed FedLPS adopt the same training hyper-parameters. In cases where explicit details are not provided, the detailed configurations of these parameters are illustrated in Table~\ref{apd_hypara_val}.

\begin{table}[!ht]
\centering\begin{tabular}{l|c}
\toprule
Hyper-parameter  & Value \\
\midrule
Layer number $n$ of the encoder in FedLPS   & $\lceil 0.25N \rceil$ \\
Learning rate       & 0.001     \\
Weight decay        & 0.001   \\
Batch size          & 512    \\
Local epoch         & 5     \\
Communication round & 100   \\
Client participation rate  & 100\%      \\
Client number       & \makecell{10}    \\
Concentration parameter $\alpha$     & \makecell{0.5}    \\
\bottomrule
\end{tabular}
\caption{Hyper-parameters used in FL training.}\label{apd_hypara_val}
\end{table}

\section{Comparison of the communication overhead}\label{apd_com_overhead}

In FL, the communication overhead is predominantly influenced by the model footprint. Consequently, we present a comparative analysis of communication overhead in Table~\ref{tab_com_overhead} to evaluate the model footprints employed across various FL frameworks. It is noteworthy that, despite FedDrop achieving the utmost reduction in model footprint, it concurrently experiences the most substantial degradation in accuracy.

\newpage
\begin{table}[!h]
\setlength\tabcolsep{10pt}
\centering\begin{tabular}{l|c|c}
\toprule
{FL frameworks} & {\makecell{ShuffleNetV2}}  & {ResNet18} \\
\midrule
FedAvg          &  6.32M  &   55.91M   \\
FedDrop         &  \textbf{1.26M}   &  \textbf{11.18M}   \\
FedProx         &  6.32M  &   55.91M  \\
Hermes          &  3.79M  &   33.54M  \\
FedMP           &  3.79M  &   33.54M   \\
\midrule
FedLPS($n = \lceil 0.25N \rceil$) &  3.73M  &  33.23M \\
FedLPS($n = \lceil 0.5N \rceil$) &  3.24M  &  32.18M \\
FedLPS($n = \lceil 0.75N \rceil$) &  \underline{2.24M}  &  \underline{27.98M} \\
\bottomrule
\end{tabular}
\caption{Comparison of the communication overhead (total transmitted parameters by all clients per FL round).}\label{tab_com_overhead}
\end{table}

% \flushend
\section{Effect of pruning ratio on the predictors}\label{apd_para_reduction}

In FedLPS, diverse pruning ratios are employed within heterogeneous federated learning to generate varying model footprints. The detailed footprints of the predictors in FedLPS are presented in Table~\ref{tab_para_reduction}.

\begin{table}[!ht]
\setlength\tabcolsep{5pt}
\centering\begin{tabular}{c|c|c|c|c}
\toprule
\multirow{2}{*}{\makecell{Pruning\\ratios $\rho$}} & \multicolumn{2}{c|}{ShuffleNetV2}  & \multicolumn{2}{c}{ResNet18} \\\cmidrule(lr){2-3}\cmidrule(lr){4-5}
    &   \#Param.  & Reduction  &   \#Param.  & Reduction \\
\midrule

0.2          &  1.02M  &   -19.5\%   &  8.98M  &   -19.7\%    \\
0.4          &  0.77M  &   -39.0\%   &  6.77M  &   -39.4\%    \\
0.6          &  0.52M  &   -58.5\%   &  4.57M  &   -59.2\%    \\
0.8          &  0.28M  &   -78.0\%   &  2.63M  &   -78.9\%    \\
\bottomrule
\end{tabular}
\caption{Parameter number of predictors (denoted by ``\#Param.'') in FedLPS ($n = \lceil 0.25N \rceil$).}\label{tab_para_reduction}
\end{table}

\end{appendices}

\end{document}